# Pruning the Path to Optimal Care: Identifying Systematically Suboptimal Medical Decision-Making with Inverse Reinforcement Learning

**Inko Bovenzi\*, Adi Carmel\*, Michael Hu\*, Rebecca Hurwitz\*, Fiona McBride\*, Leo Benac, José Roberto Tello Ayala, Finale Doshi-Velez**
**Harvard University, Cambridge, MA, USA**

\*These authors contributed equally to this work. Authorship is listed in alphabetical order.

**Abstract**

*In aims to uncover insights into medical decision-making embedded within observational data from clinical settings, we present a novel application of Inverse Reinforcement Learning (IRL) that identifies suboptimal clinician actions based on the actions of their peers. This approach centers two stages of IRL with an intermediate step to prune trajectories displaying behavior that deviates significantly from the consensus. This enables us to effectively identify clinical priorities and values from ICU data containing both optimal and suboptimal clinician decisions. We observe that the benefits of removing suboptimal actions vary by disease and differentially impact certain demographic groups.*

## 1  Introduction

Observational data from clinicians treating patients provides a valuable opportunity to understand the priorities and values guiding their decision-making. These insights can enhance clinical decision-making and guide the development of human-aligned AI decision support systems.

However, extracting clinician priorities from data is challenging. Clinicians exhibit diverse behaviors due to distinct medical backgrounds, experience levels, patient contexts, and available resources[1]. Identifying high and low-quality actions cannot be distinguished solely by examining outcomes, as patients may be non-randomly assigned to clinicians. Researchers have conducted qualitative studies such as interviews and surveys to directly inquire about clinicians' decision-making processes and values[2], and used regression analyses to identify factors that may influence clinical decision-making[3]. While these methods provide valuable insights, they fall short of capturing the nuanced and implicit aspects of decision-making.

By seeking to extract the implicit rewards that underlie observations of expert behavior, IRL offers a promising avenue to gain a more nuanced understanding of clinician decision-making[4]. For example, when treating hypotension in the ICU, an IRL method may identify that clinicians care about raising blood pressure *and* preventing fluid from accumulating in the lungs. However, IRL methods assume that the observed behavior is near-optimal; learning from data that includes suboptimal behaviors can lead to the identification of incorrect implicit reward structures, rendering them unusable for further insights.

In this work, we use IRL as a novel approach to identify suboptimal clinician decisions. First, we employ IRL to learn what clinicians prioritize based on their observed behavior. Then, using this learned standard, we identify instances where decisions deviate from the consensus reward—situations where most clinicians would agree that a better treatment choice was available given the patient's condition. This approach helps us avoid conflating the patient's condition with outcomes. By doing so, we can retrain the IRL model with a cleaner set of optimal trajectories and analyze the supposedly suboptimal ones for further insights.

We apply our approach to two decision-making settings in critical care: treating hypotension and treating sepsis. Our method effectively removes noise caused by sub-optimal behavior from the learned implicit rewards and the associated treatment policies. When we measure the reward gained from the new treatment policies, which exclude suboptimal actions, we observe a uniform increase in rewards for hypotension patients. However, for patients with sepsis, the impact is uneven. Specifically, Black patients experience a significantly higher increase in reward compared to White patients. Similar trends are observed among patients on Medicare versus Medicaid and across various marital statuses. This suggests that while identifying and correcting suboptimal treatments generally benefits all patients, it may have a particularly positive impact on certain demographic groups.

## 2 Related Work

**Evaluating Decision-Making in Healthcare Institutions** When evaluating the quality of healthcare provided, most work has been conducted on the level of healthcare institutions, measuring the impact of qualitative features such as patient satisfaction, or focusing on the relationship between treatment and cost[5]. A primary limitation of this work is that it relies on an initial benchmark for "good" and "bad" decisions, and may not fully capture the complexity and variability of medical decision-making in dynamic and uncertain environments. Additionally, a clinician's effectiveness cannot be solely judged by their outcomes, as outcomes alone do not provide insight into the aspects of care to which a particular result can be attributed[6]. Our application of IRL helps resolve this issue by distinguishing good and bad decisions irrespective of overall outcomes or subjective influences from having to pre-define quality cutoffs.

There is also work that shows disparities in treatments and outcomes across different demographic groups[7] and the impact of investigating demographic variables in healthcare algorithms[8], highlighting the importance of these considerations in creating ethical and equitable clinical decision-making support tools. Our work expands on this by providing insight into how clinician behavior varies across demographic variables and contributes to disparities between groups, and quantifies the effect of these underlying biases.

**Using IRL to Identify Expert Priorities** Reinforcement learning (RL) has already been shown to have successful applications in multiple settings involving complex sequential decision-making, including within the healthcare domain[9]. However, RL faces several challenges that limit its ability to solve dynamic problems in real healthcare settings[10,11]. IRL has the potential to solve some of the limitations of RL because it does not require an explicitly defined reward function, though it has still shown limited efficacy in providing new treatment policy recommendations[12,13]. One notable limitation of previous work applying IRL to healthcare problems is the heterogeneity of patient data and clinician decision-making[14]. Other work provides a more comprehensive review of IRL in healthcare and its limitations[15].

Our research expands on this body of work by offering a novel approach to mitigating systematic errors in expert behavior, leveraging IRL to refine decision-making models for clinical environments. Using IRL, we present a step towards resolving the problem of heterogeneity in the expert data and investigate the effects of demographic variables in the clinician decision-making process.

## 3 Background and Notation

The standard formulation for an IRL problem centers around a stationary Markov Decision Process (MDP) comprised of four elements: states, actions, transition probabilities, and rewards. States ($S$) represent conceivable situations or conditions within the decision-making context, while actions ($A$) encompass feasible moves or decisions associated with each state. Transition probabilities ($T$) articulate the probabilities linked to transitioning from one state to another, expressed as $\delta : S \times A \times S' \to [0, 1]$. Rewards ($R$) include rewards assigned for transitioning from one state to another, denoted as $R : S \to [-1, 1]$. A trajectory refers to a sequence of states and actions that describe the path that a decision-maker, such as a clinician, takes over a series of interactions within the defined MDP. We define $T$ to represent the set of trajectories in a patient cohort, where each trajectory $t$ is a sequence of state-action-next state triples $(s, a, s')$.

In the medical context, states can be characterized by vital signs. Notably, by clustering patient statuses based on their vitals data, it is possible to create a discrete set of states to define an MDP. This approach provides a data-driven method for defining states based on the patient's physiological status. Actions, in turn, can be defined in the medical setting as treatments applied to patients. Given temporally aligned vitals and treatment data, we can map this data to trajectories that can be used to train a reward model.

Several algorithms can be employed to solve an IRL problem within the MDP framework. One common technique is Maximum Entropy IRL (MaxEnt IRL). MaxEnt IRL introduces the concept of entropy to model the uncertainty in the expert's decision-making process[16]. This method aims to find a reward function that not only explains the observed behavior but also maintains a certain level of unpredictability, enhancing the interpretability and generalizability of learned reward functions.

The MaxEnt IRL optimization problem can be described as:

$$\max_R H(p(\tau|R)) - \lambda(E_{p(\tau|R)}[\phi] - E_{p^*(\tau)}[\phi])$$

where:  $p^*(\tau)$ represents empirical distribution of observed trajectories

$\lambda$ is a regularization parameter

$E_{p(\tau|R)}[\phi]$ denotes expected feature counts under trajectory distribution induced by $R$

$E_{p^*(\tau)}[\phi]$ represents expected feature counts of observed behavior

MaxEnt IRL proves particularly suitable for medical settings due to its ability to handle complex and dynamic environments. By incorporating entropy, MaxEnt IRL captures the nuanced decision-making strategies of clinicians, allowing for a more accurate representation of the underlying reward structure.

## 4 Cohort and Data Pre-Processing

The data used in this paper is a subset of MIMIC-IV, which contains detailed EHR information about ICU admissions records collected between 2008 and 2019 at Beth Israel Deaconess Medical Center in Boston, MA[17]. To make the two datasets, we extracted patient records with ICD codes related to hypotension and sepsis.

For each patient, we linked demographic information related to race and ethnicity, primary language spoken, insurance, and marital status. To mitigate the effects of small sample sizes resulting from granular race and ethnicity information in the data, we grouped patients according to the categories on the 2020 US Census (White, Black or African American, Asian, American Indian or Alaska Native, and Native Hawaiian or Other Pacific Islander). Any patients who identified as Hispanic or Latino were put in a separate category, regardless of their primary race identification. We excluded American Indian or Alaska Native, and Native Hawaiian or Other Pacific Islander people as they made up less than 1% of the data. This resulted in final datasets of 4488 hypotension patients and 6224 sepsis patients (Appendix 1).

A subset of patient vitals was extracted from the MIMIC-IV database as well as treatment information. These variables were used to create the state and action spaces. To account for missing vital data, we performed two steps of imputation. First, for each timestamp missing a vital measurement, we propagated the last known value of the vital in the patient data until the next time the vital was measured. In cases where the vital metric had not yet been measured during a given hospital stay, we used a table of 'normal' values of the vital, based on physician recommendation, to impute the patient's measurements until the first measurement of the vital was taken.

**State-Space and Action Definitions** The state space for each dataset was made up of variables related to patient vitals and labs, both of which offer important information about which states may be more desirable. The specific variables included can be found in Tables 1 and 2. These measurements are taken routinely during a hospital stay, making them accessible and informative throughout treatment. We removed extreme outlier values from both datasets and removed clusters with fewer than 10 representatives. Without this correction, we found that IRL tended to focus on signals from unreasonable states. Since MaxEnt IRL requires a discretized state space but the measured variables are continuous, we used k-means clustering to establish a discrete set of spaces $S$ such that |S| = 200 for both datasets.

We defined the action space for each condition based on the primary treatments used. As a result, the hypotension action space $A$ consists of no treatment, vasopressors, bolus epinephrine, and vasopressors + bolus epinephrine, where |A| = 4, and the sepsis action space consists of no treatment, ventilation, glucocorticoids, antibiotics, and vasoactive drugs, thus |A| = 5.

## 5 Methods

We applied MaxEnt IRL to discern the reward function from clinicians' decision-making patterns in hypotension and sepsis treatment. For each condition, hyperparameters were fine-tuned through extensive testing for convergence. For hypotension, we standardized all state feature weights to 1.0 and used an Exponential Stochastic Gradient Ascent

(ExpSGA) optimizer with a linearly decaying learning rate starting at 0.2. In contrast, for sepsis, initial state feature weights were randomly assigned from a normal distribution to encourage exploration in early optimization phases. Here, a Stochastic Gradient Ascent (SGA) optimizer with a similarly linearly decaying learning rate from 0.2 was utilized.

This initial round of MaxEnt IRL provides a preliminary reward model that infers priorities and preferences from observed expert behavior. To identify trajectories influenced by suboptimal decision-making, we present the mean expected reward loss between the optimal actions (as determined by our learned reward function, after pruning out trajectories with suboptimal behavior) and the selected actions in each trajectory:

$$L = \frac{1}{n} \sum_{t=1}^{n} E(s_t, \Pi^*(s_t)) - E(s_t, a_t); \qquad E(s, a) = \sum_{s' \in S} R(s') \cdot P(s, a, s').$$

Trajectories with more negative mean reward loss indicate deviations from the inferred optimal expert behavior that cause worse outcomes for patients.

To identify and prune out suboptimal trajectories based on mean reward loss, we considered two methodologies: Likelihood-Based Pruning and Deviation-Based Pruning. Likelihood-Based Pruning involves evaluating the likelihood of a given trajectory under the optimal reward structure and its associated policy. Policies with a lower likelihood, under the assumption that the IRL rewards are optimal, are discounted. However, this method exhibited limitations in effectively handling deviations from optimal behavior in dynamic medical settings (Appendix 2).

Consequently, we used Deviation-Based Pruning, which assesses the extent of trajectory deviation from the policy derived from the initial MaxEnt IRL iteration. This method penalizes trajectories based on the geometric mean of the exponential loss of rewards, allowing for a more comprehensive measure of optimality. Experts are not penalized based on outcomes, but rather on the expected value of their actions compared to the expected value of the "optimal" action. This method produced consistent and reliable outputs in refining the dataset for subsequent iterations.

This is formalized as follows. First, we calculate the expected reward of the optimal action $a_1$ as defined by the first iteration of IRL:

$$r_{opt} = \sum_{s} p_t(a_1, s) \cdot r_s$$

Then, we calculate the expected reward of the selected action in the same way:

$$r_{sel} = \sum_{s} p_t(a_2, s) \cdot r_s$$

Given these two rewards, we penalize actions according to their deviation from optimal rewards by taking the geometric mean of the exponential loss of rewards over the trajectory $T$ with steps $t$, calculating a score $C$.

$$C = \left( \prod_{t=1}^{n} e^{r_{sel}(t) - r_{opt}(t)} \right)^{1/n}$$

Given a score $C$, we can rank the trajectories by their optimality and then filter a certain percentile of them. In the pruning step, a proportion of trajectories with the highest mean reward loss are systematically excluded. This step refines the dataset, focusing on trajectories that align more closely with optimal expert behavior.

Following trajectory pruning, we conduct a second iteration of MaxEnt IRL on the refined dataset. This iterative process aims to emulate learning from a more optimal expert dataset, free from the influence of suboptimal decisions. The refined reward model derived from this step provides a more accurate representation of clinician priorities by focusing on trajectories that exhibit higher congruence with optimal decision-making. *While trajectories are excluded based on containing decisions that are already considered suboptimal, the key insight of the process is that trajectories with a few known suboptimal decisions are disproportionately likely to contain other unknown suboptimal decisions over a randomly selected trajectory.*

## 6 Results

**Trajectory pruning increases performance robustly across pruning parameters.** Our study centers on a trajectory pruning algorithm influenced by the fraction of trajectories it retains. Setting this fraction to 100% essentially aligns our method with traditional IRL. We tested retention rates of 20%, 50%, and 80%, and variations in state spaces determined by clustering (Appendix 3). Though our results focus on a 50% retention rate, the observed trends persist across all tested settings.

Figures 1 and 2 show the top and bottom 25 clusters before and after pruning 50% of the trajectories for both the hypotension and sepsis data.

**Table 1:** Comparative analysis of the mean values and standard deviations for key clinical features across the best and worst clusters for hypotension data. Pruned clusters exhibit slight changes in standard deviations, with some features indicating increased variability, particularly in the worst clusters. The best pruned clusters display mean values that are generally closer to clinical norms, while the worst pruned clusters demonstrate more pronounced changes in both mean values and variability. This suggests that pruning refines the dataset by highlighting more significant deviations in the worst clusters, potentially enhancing the algorithm's ability to identify critical outliers.

| Feature | Best Clusters | | | | | | Worst Clusters | | | | | |
| --- | --- | --- | --- | --- | --- | --- | --- | --- | --- | --- | --- | --- |
| | Unpruned | | Pruned | | Difference | | Unpruned | | Pruned | | Difference | |
| | Mean | Std | Mean | Std | $\Delta$Mean | $\Delta$Std | Mean | Std | Mean | Std | $\Delta$Mean | $\Delta$Std |
| Creat. (mg/dL) | 1.42 | 0.11 | 1.44 | 0.11 | 0.02 | 0.00 | 2.07 | 0.56 | 1.91 | 0.56 | -0.16 | 0.00 |
| $FiO_2$ (%) | 0.49 | 0.008 | 0.49 | 0.008 | 0.00 | 0.000 | 0.55 | 0.079 | 0.56 | 0.097 | 0.01 | 0.018 |
| Lac. (mmol/L) | 1.95 | 0.11 | 1.98 | 0.11 | 0.03 | 0.00 | 4.53 | 2.62 | 5.30 | 4.20 | 0.77 | 1.58 |
| Urine Out. (mL) | 95.92 | 4.88 | 95.75 | 4.14 | -0.17 | -0.74 | 87.85 | 32.44 | 82.95 | 32.80 | -4.90 | 0.36 |
| ALT (U/L) | 120.9 | 14.73 | 122.5 | 15.10 | 1.62 | 0.37 | 399.7 | 423.4 | 812.8 | 1164.0 | 413.1 | 740.6 |
| AST (U/L) | 200.6 | 28.52 | 203.1 | 28.22 | 2.50 | -0.30 | 858 | 969 | 2067 | 3492 | 1209 | 2523 |
| MBP (mmHg) | 68.64 | 0.99 | 68.75 | 1.11 | 0.11 | 0.12 | 75.32 | 40.84 | 76.49 | 39.82 | 1.17 | -1.02 |
| DBP (mmHg) | 51.72 | 1.03 | 51.95 | 1.22 | 0.23 | 0.19 | 54.98 | 11.69 | 55.21 | 10.49 | 0.23 | -1.20 |
| SBP (mmHg) | 108.80 | 1.89 | 108.56 | 1.66 | -0.24 | -0.23 | 101.22 | 18.59 | 103.07 | 12.30 | 1.85 | -6.29 |
| GCS | 14.82 | 0.019 | 14.82 | 0.021 | 0.00 | 0.002 | 14.22 | 0.82 | 14.44 | 0.51 | 0.22 | -0.31 |
| $PaO_2$ (mmHg) | 121.01 | 3.81 | 120.51 | 3.88 | -0.50 | 0.07 | 130.48 | 32.96 | 119.02 | 16.56 | -11.46 | -16.40 |
| HR (BPM) | 79.31 | 7.50 | 81.33 | 7.15 | 2.02 | -0.35 | 94.79 | 59.46 | 103.05 | 65.93 | 8.26 | 6.47 |
| Temp (°C) | 36.92 | 0.066 | 36.93 | 0.059 | 0.01 | -0.007 | 36.86 | 0.53 | 36.76 | 0.36 | -0.10 | -0.17 |
| RR (Breaths/Min) | 18.13 | 2.96 | 18.55 | 3.25 | 0.42 | 0.29 | 38.06 | 56.01 | 27.19 | 23.68 | -10.87 | -32.33 |

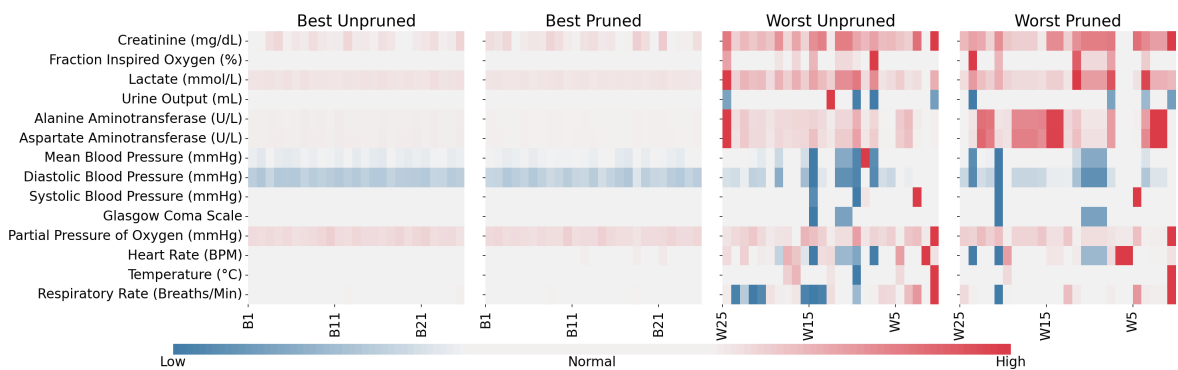

**Figure 1:** Comparison of clusters with the 25 highest and lowest rewards from unpruned and pruned MaxEntIRL runs for hypotension data. The color gradients represent the extent to which the features align with or deviate from the normal ranges. Lighter colors indicate a shift towards normality in pruned clusters, especially seen in the highest-reward clusters. Darker colors in the lowest-reward clusters reflect the more severe deviations, particularly seen in the ALT and AST levels, that pruning appears to effectively penalize.

**Table 2:** Comparative analysis of the mean values and standard deviations for key clinical features across the best and worst clusters for sepsis data. Pruned clusters exhibit lower standard deviations, indicating enhanced consistency, whereas unpruned clusters show increased deviations, indicating variability. The best pruned clusters exhibit mean values closer to expected clinical norms, whereas the best unpruned clusters demonstrate exacerbated mean values, suggesting that pruning enhances the algorithm's capacity to discern more reliable signals.

| Feature | Best Clusters | | | | | | Worst Clusters | | | | | |
|---|---|---|---|---|---|---|---|---|---|---|---|---|
| | Unpruned | | Pruned | | Difference | | Unpruned | | Pruned | | Difference | |
| | Mean | Std | Mean | Std | ΔMean | ΔStd | Mean | Std | Mean | Std | ΔMean | ΔStd |
| HR (BPM) | 90.58 | 18.38 | 85.96 | 13.95 | -4.62 | -4.43 | 81.14 | 39.19 | 93.86 | 27.31 | 12.72 | -11.88 |
| MBP (mmHg) | 75.64 | 15.58 | 72.49 | 7.83 | -3.15 | -7.75 | 88.78 | 46.88 | 100.51 | 55.47 | 11.73 | 8.59 |
| RR (Breaths/Min) | 21.68 | 5.60 | 21.55 | 4.41 | -0.13 | -1.19 | 21.78 | 8.46 | 20.46 | 10.18 | -1.32 | 1.72 |
| Temp (°C) | 36.89 | 0.65 | 36.96 | 0.64 | 0.07 | -0.01 | 36.00 | 3.67 | 35.45 | 3.88 | -0.55 | 0.21 |
| FiO$_2$ (%) | 0.97 | 0.16 | 1.00 | 0.00 | 0.03 | -0.16 | 0.91 | 0.26 | 0.94 | 0.22 | 0.03 | -0.04 |

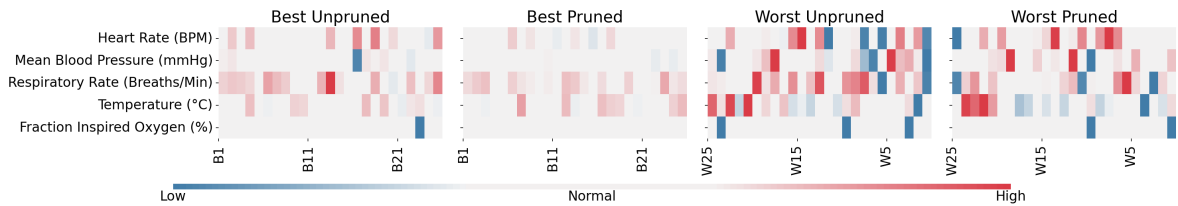

**Figure 2:** Comparison of clusters with the 25 highest and lowest rewards from unpruned and pruned MaxEntIRL runs for sepsis data. The color gradients represent the extent to which the features align with or deviate from the normal ranges. Lighter colors indicate a shift towards normality in pruned clusters.

Figure 3a shows reward changes in the hypotension data, where the states with reward changes had higher increases in reward than the others (p = 0.001), suggesting that suboptimal clinician actions hampered the algorithm's understanding of these states. Figure 3b shows reward changes in the sepsis data, where we see a much more even distribution of states with reward changes. This divergence may be due to differences in disease progression, the time-sensitivity of treatments, or other fundamental differences in how patients react to the two conditions. The high variance in sepsis rewards between iterations was consistent across multiple runs.

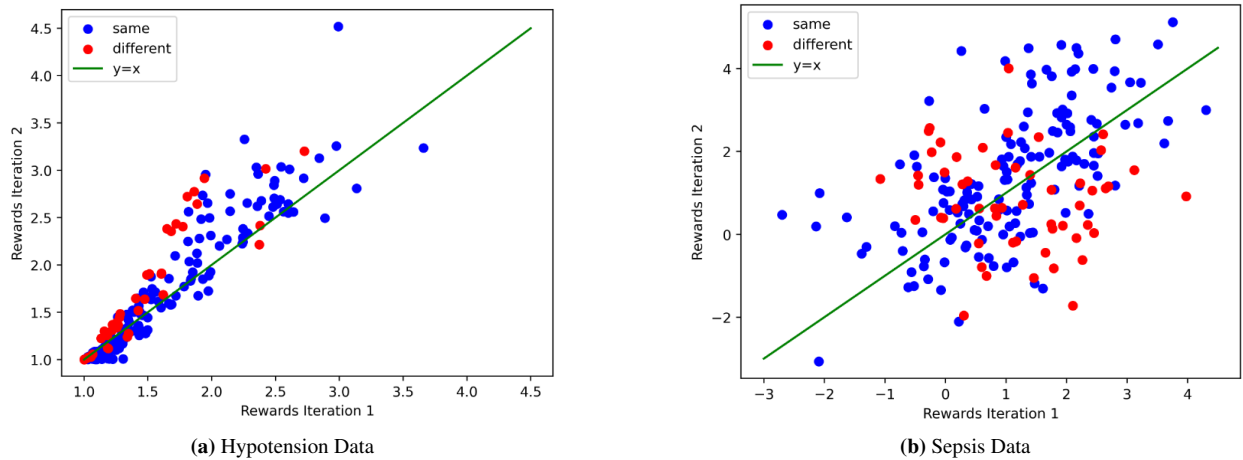

(a) Hypotension Data     (b) Sepsis Data

**Figure 3:** Reward variation by state between the first and second iterations of IRL, where the first is trained with no pruning and the second is trained with pruning. The figures illustrate changes in policy recommendations, with colors indicating agreement in policy across states. Significant policy shifts after pruning suggest that excluding suboptimal trajectories can influence the underlying reward distribution.

**End-state rewards correlate with our trajectory rankings.** Our pruning method ranks trajectories by the geometric mean of their suboptimality from the consensus action, with the hypothesis being that a trajectory with a few known suboptimal decisions may contain other suboptimal decisions. We emphasize that we did not rank trajectories based on their final end-state reward, we ranked them by the suboptimality of their previous actions. In Figure 4, trajectories that we score poorly in our algorithm (and thus drop from consideration) may be more likely to have a bad terminal state. The end-state rewards for hypotension patients level off after the bottom 20% of trajectories. The end-state rewards for sepsis patients show a general increase over the percentiles.

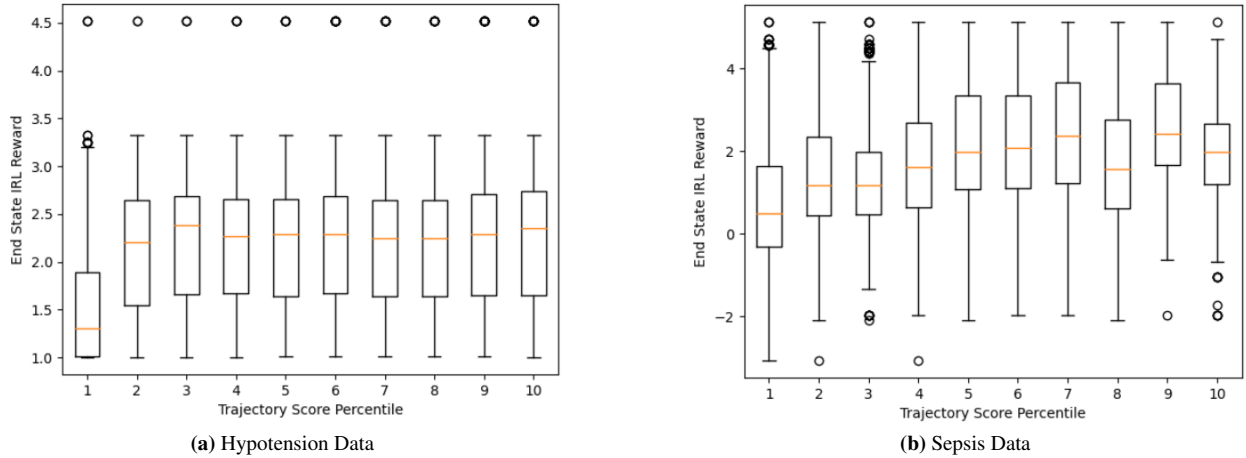

(a) Hypotension Data            (b) Sepsis Data

**Figure 4:** Trajectories are grouped into ten groups based on their percentile and average end-state reward for each group. Trajectories in the lower percentiles tend to have worse end-state rewards.

**Trajectories with suboptimal actions are spread evenly across demographics, but eliminating suboptimal trajectories differentially benefits different demographic groups.** Table 3 shows that the likelihood of a trajectory to be pruned is typically not statistically different across race, insurance type, marital status, and whether the patient is an English speaker. The only exception is that Hispanic/Latino patients are more likely to be pruned out in the sepsis data, but this does not translate into significant differences in the mean expected reward difference after pruning.

On one hand, this suggests that suboptimal actions, as identified by our IRL algorithm, are generally evenly spread across different demographic groups. However, our IRL focuses on identifying and removing trajectories of anomalous actions, where anomalous is defined according to the consensus of the clinicians in the dataset. Removing these suboptimal trajectories had a more significant impact on the rewards obtained by certain demographic groups than others. Thus, our result can still be consistent with work that exposes disparities in treatment[7,18,19].

**Table 3:** P-value significance of pruning variation among demographic groups and subsequent effects on mean expected reward difference. Almost no significant differences were seen in the proportions of pruned trajectories across demographic groups. There were significant differences in the expected reward loss for various demographic groups.

|  | **Hypotension** | | | | **Sepsis** | | | |
|---|---|---|---|---|---|---|---|---|
|  | Race | Insurance | Language | Marital Status | Race | Insurance | Language | Marital Status |
| Difference in pruning (%) | 0 | 0 | 0 | 0 | * | 0 | 0 | 0 |
| Mean expected reward loss after pruning | * | 0 | ** | *** | 0 | 0 | 0 | 0 |

\* $p < 0.05$, \*\* $p < 0.01$, \*\*\* $p < 0.001$

The absolute value of the change in reward obtained after running the IRL trained without the pruned trajectories is larger for the sepsis data than for the hypotension data. However, these changes are not significantly different among demographic groups in the sepsis data, indicating that deviating from the consensus action harms all patients equally.

While the absolute changes in reward are smaller for patients with hypotension, these changes differ significantly concerning race (ANOVA, p = 0.036), insurance (ANOVA, p = 0.007), and marital status (ANOVA, p = 1.71e-10). We performed Tukey's Honestly Significant Difference tests to investigate pairwise comparisons between the levels of each demographic group. Within race, we found a significant difference between patients identifying as Black and those identifying as White (mean expected reward loss difference = 0.0098, p = 0.026). Patients on Medicare versus Medicaid had a mean expected reward loss difference = 0.0111 (p = 0.0098). There were multiple differences seen based on marital status: divorced versus single mean diff = -0.0119 (p = 0.0021), married versus single mean diff = -0.0138 (p = 0), single versus widowed mean diff = 0.0159 (p = 0). We found no significant difference in reward loss between English and non-English speakers.

Across both datasets, the trajectories selected for pruning do not appear to correlate with one metric of patient outcomes: whether or not the patient died in the hospital (Chi-squared, hypotension p=0.29, sepsis p=0.51).

# 7 Discussion

The current landscape of IRL in healthcare research often assumes that experts consistently make optimal or near-optimal decisions, leaving a gap in understanding how IRL algorithms handle systematically suboptimal expert behavior. To address this challenge, we propose a novel two-stage approach to IRL, wherein the training data for the second iteration excludes trajectories displaying behavior that deviates significantly from the consensus based on the first iteration. This method aims to improve the overall accuracy of learned policies by excluding suboptimal trajectories, particularly in scenarios where a large proportion of experts exhibit suboptimal behavior.

The goal of our algorithm is not to change perceptions of every state; rather, it is to discard erroneous trajectories and thus change the policy in states dominated by problematic trajectories. Our study demonstrates the effectiveness of this approach by applying it to subsets of the MIMIC-IV dataset composed of patients with hypotension and sepsis. The dataset's richness and complexity make it well-suited for our algorithm, as the ICU presents numerous challenges for experts, potentially leading to systematically suboptimal trajectories. After running our algorithm on the data, we observed a small but meaningful change in the learned policies, indicating enhanced accuracy compared to traditional IRL methods. Non-identifiability is a primary concern with IRL; for any given policy, there exist infinitely many sets of rewards that yield the same policy[20]. Because we see meaningful changes in policy, however, we know that the two sets of rewards are not different instances of the same fundamental reward function. The benefit of pruning trajectories is robust across a wide range of percentages.

Interestingly, the trajectories that were pruned were not correlated with patient death for both sets of patients. While death is one outcome measure, it is not the only nor most nuanced, so the lack of correlation does not rule out a connection with other outcome measures. Measuring patient outcomes is difficult given that the starting state of patients differs, and what would be considered a good outcome may vary depending on multiple factors. Measuring a patient's return to normal vitals or the length of their hospital stay may be more informative measures of outcome, and is left to future study.

In the hypotension data, our analysis revealed interesting demographic disparities between White and Black patients. Black patients had a significantly more negative mean expected reward loss, meaning that there was a bigger difference in the reward accumulated before and after suboptimal trajectory pruning. A smaller difference in reward accumulated before and after pruning indicates that those patients are already getting the consensus treatment most of the time, so standardizing treatment would not significantly change their care. The difference in mean expected reward loss indicates that Black patients would benefit more from consensus treatment than White patients, suggesting disparities in the quality of care received. This finding is consistent with previous research highlighting racial disparities in healthcare outcomes and underscores the importance of addressing systemic biases in clinical decision-making[7,18,19].

With respect to marital status, we found that single patients were the group most likely to benefit from consensus treatment than patients with a different marital status. Previous work has demonstrated that having someone who can advocate for the patient results in improved care, which would be most expected for married patients[21]. This may explain why married patients receive better care than single patients. Single patients may have worse outcomes compared to widowed and divorced patients as a result of confounding due to age. Divorced and widowed patients are likely to be older and may receive more attention or have additional family members who can advocate for them.

Private insurance plans are not captured in the MIMIC-IV data, so our analyses related to insurance are limited. We found that hypotension patients on Medicaid would benefit more from consensus treatment than patients on Medicare. This may be due to confounding factors such as increased age and additional disabilities in Medicare patients.

We were unable to recreate the same results in the sepsis dataset, indicating the presence of specific factors that make certain diseases more likely to be influenced by demographic variables. One contributing factor may be that sepsis has more clearly defined treatment guidelines, primarily the administration of antibiotics[22]. Because clinicians need to act quickly and there is less subjectivity in treatment, sepsis may not be as susceptible to implicit biases and confounding variables. On the other hand, many ICUs do not have specific hypotension treatment protocols, so clinicians may be more subjective in their decision-making, uncovering more biases[23].

For hypotension patients in states where the reward changed after pruning, the new reward was often higher, indicating these patients benefited more from the consensus treatment. The random distribution of reward changes among the sepsis data suggests fewer confounding factors affecting the rewards. This is supported by the patterns seen in end-state reward across trajectory percentiles. Increases in end-state reward may be due to physicians making better treatment decisions, irrespective of demographic features, as seen in the sepsis data. In the hypotension data, end-state rewards level off quickly after the bottom 20%, and demographic variables play a larger role in treatment decisions.

A limitation of our approach is the assumption of a clear consensus on what constitutes optimal behavior among clinicians. In reality, clinician behavior varies significantly due to differences in medical training, experience, patient demographics, and resource availability. IRL is a tool for understanding consensus and making recommendations, but this can lead to oversimplified recommendations that do not account for the full complexity of clinical decision-making. Future research should develop methods to better incorporate this variability and include a broader range of clinical opinions to more accurately capture the diversity in medical decision-making.

Our approach is based on an initial IRL model trained on all trajectories, including those deemed suboptimal. While this method provides significant benefits, it naturally introduces bias in identifying suboptimal trajectories. If the initial model is inaccurate, relevant data may be pruned, misrepresenting clinician behavior. To mitigate this risk, future work could employ multiple random initializations of the IRL process and consistently remove only those trajectories that rank in the bottom fraction of total rewards. Implementing additional rounds of IRL and pruning, with a smaller fraction of trajectories being removed in each iteration, could also help reduce the bias introduced by the initial model.

The application of IRL in clinical decision-making requires several ethical considerations. A key concern is the potential impact on clinician autonomy and patient trust. To address this, it is essential to educate clinicians on how to use IRL systems as decision aids while ensuring that they still draw from their own experience and the circumstances of each case to make final decisions. This approach helps maintain clinician autonomy and ensures that patients can trust their healthcare providers, knowing that ultimate decisions are informed by both data-driven analysis and clinical expertise. Addressing potential biases in IRL systems is also critical. Our findings highlight demographic disparities in treatment outcomes, underscoring the need for continuous monitoring and thoughtful application of IRL algorithms. Implementing regular evaluations and incorporating diverse datasets can help ensure these systems are fair and effective.

By addressing these research avenues, we can advance our understanding of how IRL can improve clinical decision-making, enhance patient outcomes, and promote health equity in healthcare delivery.

**Acknowledgements** This material is based upon work supported by the National Science Foundation under Grant Nos. IIS-2007076 and IIS-2107391. Any opinions, findings, and conclusions or recommendations expressed in this material are those of the author(s) and do not necessarily reflect the views of the National Science Foundation. It is also supported by NIH award 5R01MH123804-02 and Harvard ERC training grant T42 OH008416.

**Appendix 1**

Self-reported demographic breakdown of patients included in this study.

| Demographic | Sepsis | Hypotension |
|---|---|---|
| White | 4929 | 3843 |
| Black | 812 | 349 |
| Asian | 264 | 174 |
| Hispanic/Latino | 219 | 122 |
| Medicare | 3331 | 2044 |
| Medicaid | 439 | 246 |
| Other | 2454 | 2198 |
| English | 5583 | 4160 |
| Other/Unknown | 641 | 328 |
| Married | 2827 | 2500 |
| Single | 1895 | 1102 |
| Divorced | 484 | 413 |
| Widowed | 918 | 398 |

**Appendix 2**

Two options for determining which trajectories to prune include Likelihood-Based Pruning and Deviation-Based Pruning. We chose to use Deviation-Based Pruning, as it was better suited to the complex medical data. However, we describe the Likelihood-Based Pruning algorithm below as an alternate method that may be effective on other types of data:

Let $T$ be a set of MIMIC trajectories, where each trajectory $t$ is a sequence of state-action-next state triples $(s, a, s')$. Let $\pi$ represent the policy, where $\pi(s)$ gives the action to take in state $s$, and $P$ represents the transition probabilities, where $P(s, s', a)$ gives the probability of transitioning to state $s'$ from state $s$ when action $a$ is taken.

The likelihood $L$ of a trajectory $t$ under the policy $\pi$ and transition probabilities $P$ is given by:

$$L(t|\pi, P) = \prod_{(s,a,s') \in t} P(s, s', a)^{I(\pi(s)=a)}$$

where $I(\pi(s) = a)$ is an indicator function that equals 1 when the action $a$ is the one recommended by the policy $\pi$ for state $s$, and 0 otherwise.

Thus, the log-likelihood $\ell$ of trajectory $t$ is the logarithm of $L$:

$$\ell(t|\pi, P) = \sum_{(s,a,s') \in t} \log P(s, s', a) \cdot I(\pi(s) = a)$$

The set of pruned trajectories $T'$ based on a percentile $p$ or threshold $\theta$ is then:

$$T' = \{t \in T | \ell(t|\pi, P) \geq c\}$$

where $c$ is a cutoff log-likelihood, defined as:

1. If percentile $p$ is given: $c = \text{percentile}(\ell(T|\pi, P), 100 - p)$
2. If threshold $\theta$ is given: $c = \log(\theta)$

We keep all trajectories whose log-likelihood is greater than or equal to the cutoff $c$. The percentile function computes the value below which a given percentage of observations in a set of values fall, thus filtering out the lower-percentage likelihood trajectories.

## Appendix 3

The tables below display the mean values and standard deviations of key clinical features and rewards across different pruning thresholds (20% and 80%) for both ranked pruning (using the deviation metric) and one iteration of random pruning (removing trajectories at random) in data from hypotension patients. Values for random pruning change with every iteration, depending on the trajectories selected for removal.

For ranked pruning, we observe that as the pruning threshold increases, the mean values of clinical features approach more optimal, clinically desirable ranges. This trend suggests that higher pruning thresholds more effectively filter out suboptimal trajectories, resulting in clusters that better reflect high-quality decision-making. Additionally, the standard deviations decrease with higher pruning thresholds in ranked pruning, indicating reduced variability and a more homogeneous set of near-optimal trajectories.

In contrast, with random pruning, there is no clear trend toward optimal mean values, and the standard deviations remain relatively high across thresholds, reflecting greater variability and inconsistency in clinical outcomes.

**Table 4:** Comparative analysis of the mean values and standard deviations for key clinical features across the best and worst clusters for threshold 20%.

| Feature | Best Clusters | | | | | | Worst Clusters | | | | | |
| --- | --- | --- | --- | --- | --- | --- | --- | --- | --- | --- | --- | --- |
| | Unpruned | | Pruned | | Difference | | Unpruned | | Pruned | | Difference | |
| | Mean | Std | Mean | Std | ΔMean | ΔStd | Mean | Std | Mean | Std | ΔMean | ΔStd |
| Creatinine (mg/dL) | 1.42 | 0.11 | 1.57 | 0.14 | 0.15 | 0.03 | 2.14 | 0.68 | 2.51 | 0.14 | 0.37 | -0.53 |
| Fraction Inspired Oxygen (%) | 0.49 | 0.01 | 0.49 | 0.01 | 0.00 | 0.00 | 0.54 | 0.09 | 0.62 | 0.04 | 0.07 | -0.06 |
| Lactate (mmol/L) | 1.92 | 0.10 | 2.13 | 0.15 | 0.22 | 0.05 | 4.32 | 2.50 | 7.37 | 2.50 | 3.05 | 0.00 |
| Urine Output (mL) | 96.57 | 5.79 | 105.33 | 5.73 | 8.76 | -0.06 | 85.82 | 37.10 | 71.43 | 21.86 | -14.39 | -15.24 |
| Alanine Aminotransferase (U/L) | 119.00 | 16.27 | 140.40 | 15.00 | 21.40 | -1.28 | 308.05 | 241.75 | 827.42 | 934.94 | 519.37 | 693.19 |
| Aspartate Aminotransferase (U/L) | 195.34 | 28.89 | 240.58 | 30.01 | 45.24 | 1.12 | 709.80 | 624.66 | 2186.70 | 2179.28 | 1476.89 | 1554.63 |
| Mean Blood Pressure (mmHg) | 68.63 | 1.00 | 71.17 | 1.39 | 2.54 | 0.39 | 78.37 | 53.06 | 47.14 | 12.90 | -31.23 | -40.16 |
| Diastolic Blood Pressure (mmHg) | 51.58 | 1.05 | 55.17 | 1.33 | 3.58 | 0.29 | 54.52 | 14.44 | 39.43 | 7.83 | -15.09 | -6.61 |
| Systolic Blood Pressure (mmHg) | 109.10 | 2.09 | 108.12 | 1.82 | -0.99 | -0.27 | 98.81 | 23.58 | 76.27 | 18.01 | -22.54 | -5.56 |
| Glasgow Coma Scale | 14.84 | 0.01 | 14.71 | 0.07 | -0.13 | 0.06 | 14.03 | 1.00 | 12.66 | 1.02 | -1.37 | 0.02 |
| Partial Pressure of Oxygen (mmHg) | 121.02 | 3.86 | 120.26 | 5.55 | -0.76 | 1.69 | 136.55 | 40.03 | 130.09 | 16.12 | -6.46 | -23.91 |
| Heart Rate (BPM) | 77.46 | 7.97 | 102.61 | 5.89 | 25.15 | -2.08 | 87.69 | 71.29 | 33.23 | 39.35 | -54.46 | -31.94 |
| Temperature (°C) | 36.90 | 0.06 | 37.06 | 0.04 | 0.16 | -0.02 | 36.79 | 0.62 | 36.45 | 0.26 | -0.34 | -0.36 |
| Respiratory Rate (Breaths/Min) | 18.25 | 2.50 | 18.99 | 3.34 | 0.74 | 0.84 | 52.14 | 68.40 | 13.84 | 11.74 | -38.30 | -56.65 |
| Rewards | 0.77 | 0.17 | 0.73 | 0.26 | -0.04 | 0.10 | -0.81 | 0.01 | -0.80 | 0.00 | 0.02 | -0.01 |

**Table 5:** Comparative analysis of the mean values and standard deviations for key clinical features across the best and worst clusters for threshold 80%.

| Feature | Best Clusters | | | | | | Worst Clusters | | | | | |
|---|---|---|---|---|---|---|---|---|---|---|---|---|
| | Unpruned | | Pruned | | Difference | | Unpruned | | Pruned | | Difference | |
| | Mean | Std | Mean | Std | ΔMean | ΔStd | Mean | Std | Mean | Std | ΔMean | ΔStd |
| Creatinine (mg/dL) | 1.42 | 0.11 | 1.42 | 0.12 | -0.00 | 0.01 | 2.14 | 0.68 | 2.31 | 0.66 | 0.17 | -0.02 |
| Fraction Inspired Oxygen (%) | 0.49 | 0.01 | 0.49 | 0.01 | 0.00 | -0.00 | 0.54 | 0.09 | 0.54 | 0.10 | 0.00 | 0.00 |
| Lactate (mmol/L) | 1.92 | 0.10 | 1.93 | 0.11 | 0.01 | 0.00 | 4.32 | 2.50 | 6.49 | 4.89 | 2.17 | 2.39 |
| Urine Output (mL) | 96.57 | 5.79 | 96.60 | 5.74 | 0.04 | -0.05 | 85.82 | 37.10 | 67.99 | 26.44 | -17.83 | -10.66 |
| Alanine Aminotransferase (U/L) | 119.00 | 16.27 | 119.69 | 16.07 | 0.68 | -0.21 | 308.05 | 241.75 | 1511.66 | 1638.34 | 1203.62 | 1396.59 |
| Aspartate Aminotransferase (U/L) | 195.34 | 28.89 | 196.81 | 27.67 | 1.47 | -1.22 | 709.80 | 624.66 | 3621.53 | 3827.10 | 2911.73 | 3202.45 |
| Mean Blood Pressure (mmHg) | 68.63 | 1.00 | 68.79 | 1.02 | 0.16 | 0.02 | 78.37 | 53.06 | 70.90 | 14.92 | -7.47 | -38.15 |
| Diastolic Blood Pressure (mmHg) | 51.58 | 1.05 | 51.78 | 1.07 | 0.20 | 0.02 | 54.52 | 14.44 | 56.34 | 11.52 | 1.83 | -2.92 |
| Systolic Blood Pressure (mmHg) | 109.10 | 2.09 | 109.11 | 2.14 | 0.01 | 0.06 | 98.81 | 23.58 | 105.79 | 18.71 | 6.98 | -4.87 |
| Glasgow Coma Scale | 14.84 | 0.01 | 14.83 | 0.02 | -0.01 | 0.01 | 14.03 | 1.00 | 14.36 | 0.91 | 0.33 | -0.09 |
| Partial Pressure of Oxygen (mmHg) | 121.02 | 3.86 | 120.70 | 3.78 | -0.32 | -0.09 | 136.55 | 40.03 | 130.35 | 40.54 | -6.21 | 0.52 |
| Heart Rate (BPM) | 77.46 | 7.97 | 78.39 | 8.47 | 0.94 | 0.50 | 87.69 | 71.29 | 104.99 | 60.55 | 17.30 | -10.74 |
| Temperature (°C) | 36.90 | 0.06 | 36.91 | 0.07 | 0.00 | 0.00 | 36.79 | 0.62 | 36.81 | 0.56 | 0.02 | -0.06 |
| Respiratory Rate (Breaths/Min) | 18.25 | 2.50 | 18.20 | 2.59 | -0.06 | 0.09 | 52.14 | 68.40 | 58.94 | 70.55 | 6.80 | 2.15 |
| Rewards | 0.77 | 0.17 | 0.77 | 0.17 | 0.01 | 0.00 | -0.81 | 0.01 | -0.81 | 0.00 | -0.00 | -0.01 |

**Table 6:** Comparative analysis of the mean values and standard deviations for key clinical features across the best and worst clusters for one iteration of random pruning by half. The values for the "best" and "worst" clusters change with every new iteration of random pruning.

| Feature | Best Clusters | | | | | | Worst Clusters | | | | | |
|---|---|---|---|---|---|---|---|---|---|---|---|---|
| | Unpruned | | Pruned | | Difference | | Unpruned | | Pruned | | Difference | |
| | Mean | Std | Mean | Std | ΔMean | ΔStd | Mean | Std | Mean | Std | ΔMean | ΔStd |
| Creatinine (mg/dL) | 1.42 | 0.11 | 1.42 | 0.12 | 0.00 | 0.01 | 2.14 | 0.68 | 2.11 | 0.69 | -0.03 | 0.01 |
| Fraction Inspired Oxygen (%) | 0.49 | 0.01 | 0.49 | 0.01 | 0.00 | 0.00 | 0.54 | 0.09 | 0.54 | 0.09 | 0.00 | 0.00 |
| Lactate (mmol/L) | 1.92 | 0.10 | 1.92 | 0.10 | 0.00 | 0.00 | 4.32 | 2.50 | 4.35 | 2.48 | 0.03 | -0.02 |
| Urine Output (mL) | 96.57 | 5.79 | 96.99 | 5.84 | 0.42 | 0.05 | 85.82 | 37.10 | 84.28 | 36.90 | -1.54 | -0.20 |
| Alanine Aminotransferase (U/L) | 119.00 | 16.27 | 119.43 | 16.12 | 0.43 | -0.15 | 308.05 | 241.75 | 325.65 | 235.35 | 17.60 | -6.40 |
| Aspartate Aminotransferase (U/L) | 195.34 | 28.89 | 198.58 | 27.59 | 3.24 | -1.30 | 709.80 | 624.66 | 734.46 | 615.86 | 24.66 | -8.80 |
| Mean Blood Pressure (mmHg) | 68.63 | 1.00 | 68.68 | 0.96 | 0.05 | -0.04 | 78.37 | 53.06 | 65.64 | 18.79 | -12.73 | -34.27 |
| Diastolic Blood Pressure (mmHg) | 51.58 | 1.05 | 51.64 | 1.03 | 0.06 | -0.02 | 54.52 | 14.44 | 53.32 | 14.05 | -1.20 | -0.39 |
| Systolic Blood Pressure (mmHg) | 109.10 | 2.09 | 109.10 | 2.08 | 0.00 | -0.01 | 98.81 | 23.58 | 97.26 | 22.68 | -1.55 | -0.90 |
| Glasgow Coma Scale | 14.84 | 0.01 | 14.83 | 0.01 | -0.01 | 0.00 | 14.03 | 1.00 | 14.05 | 1.01 | 0.02 | 0.01 |
| Partial Pressure of Oxygen (mmHg) | 121.02 | 3.86 | 121.46 | 4.03 | 0.44 | 0.17 | 136.55 | 40.03 | 138.67 | 39.93 | 2.12 | -0.10 |
| Heart Rate (BPM) | 77.46 | 7.97 | 77.50 | 8.00 | 0.04 | 0.03 | 87.69 | 71.29 | 86.90 | 71.65 | -0.79 | 0.36 |
| Temperature (°C) | 36.90 | 0.06 | 36.90 | 0.06 | 0.00 | 0.00 | 36.79 | 0.62 | 36.78 | 0.62 | -0.01 | 0.00 |
| Respiratory Rate (Breaths/Min) | 18.25 | 2.50 | 18.03 | 2.77 | -0.22 | 0.27 | 52.14 | 68.40 | 50.96 | 69.28 | -1.18 | 0.88 |
| Rewards | 0.77 | 0.17 | 0.76 | 0.17 | -0.01 | 0.00 | -0.81 | 0.01 | -0.81 | 0.01 | 0.00 | -0.00 |

**Table 7:** Comparative analysis of the changes in mean values and standard deviations for key clinical features across the best clusters for thresholds 20%, 50%, 80%, and random pruning by half.

| Feature | 20% ΔMean | 20% ΔStd | 50% ΔMean | 50% ΔStd | 80% ΔMean | 80% ΔStd | Random ΔMean | Random ΔStd |
|---|---|---|---|---|---|---|---|---|
| Creatinine (mg/dL) | 0.15 | 0.03 | 0.02 | 0.00 | -0.00 | 0.01 | 0.00 | 0.01 |
| Fraction Inspired Oxygen (%) | 0.00 | 0.00 | 0.00 | 0.00 | 0.00 | -0.00 | 0.00 | 0.00 |
| Lactate (mmol/L) | 0.22 | 0.05 | 0.03 | 0.00 | 0.01 | 0.00 | 0.00 | 0.00 |
| Urine Output (mL) | 8.76 | -0.06 | -0.17 | -0.74 | 0.04 | -0.05 | 0.42 | 0.05 |
| Alanine Aminotransferase (U/L) | 21.40 | -1.28 | 1.62 | 0.37 | 0.68 | -0.21 | 0.43 | -0.15 |
| Aspartate Aminotransferase (U/L) | 45.24 | 1.12 | 2.50 | -0.30 | 1.47 | -1.22 | 3.24 | -1.30 |
| Mean Blood Pressure (mmHg) | 2.54 | 0.39 | 0.11 | 0.12 | 0.16 | 0.02 | 0.05 | -0.04 |
| Diastolic Blood Pressure (mmHg) | 3.58 | 0.29 | 0.23 | 0.19 | 0.20 | 0.02 | 0.06 | -0.02 |
| Systolic Blood Pressure (mmHg) | -0.99 | -0.27 | -0.24 | -0.23 | 0.01 | 0.06 | 0.00 | -0.01 |
| Glasgow Coma Scale | -0.13 | 0.06 | 0.00 | 0.00 | -0.01 | 0.01 | -0.01 | 0.00 |
| Partial Pressure of Oxygen (mmHg) | -0.76 | 1.69 | -0.50 | 0.07 | -0.32 | -0.09 | 0.44 | 0.17 |
| Heart Rate (BPM) | 25.15 | -2.08 | 2.02 | -0.35 | 0.94 | 0.50 | 0.04 | 0.03 |
| Temperature (°C) | 0.16 | -0.02 | 0.01 | -0.01 | 0.00 | 0.00 | 0.00 | 0.00 |
| Respiratory Rate (Breaths/Min) | 0.74 | 0.84 | 0.42 | 0.29 | -0.06 | 0.09 | -0.22 | 0.27 |
| Rewards | -0.04 | 0.10 | 0.01 | 0.00 | 0.01 | 0.00 | -0.01 | 0.00 |

**Table 8:** Comparative analysis of the changes in mean values and standard deviations for key clinical features across the worst clusters for thresholds 20%, 50%, 80%, and random pruning by half.

| Feature | 20% ΔMean | 20% ΔStd | 50% ΔMean | 50% ΔStd | 80% ΔMean | 80% ΔStd | Random ΔMean | Random ΔStd |
|---|---|---|---|---|---|---|---|---|
| Creatinine (mg/dL) | 0.37 | -0.53 | -0.16 | 0.00 | 0.17 | -0.02 | -0.03 | 0.01 |
| Fraction Inspired Oxygen (%) | 0.07 | -0.06 | 0.01 | 0.02 | 0.00 | 0.00 | 0.00 | 0.00 |
| Lactate (mmol/L) | 3.05 | 0.00 | 0.77 | 1.58 | 2.17 | 2.39 | 0.03 | -0.02 |
| Urine Output (mL) | -14.39 | -15.24 | -4.90 | 0.36 | -17.83 | -10.66 | -1.54 | -0.20 |
| Alanine Aminotransferase (U/L) | 519.37 | 693.19 | 413.10 | 740.60 | 1203.62 | 1396.59 | 17.60 | -6.40 |
| Aspartate Aminotransferase (U/L) | 1476.89 | 1554.63 | 1209.00 | 2523.00 | 2911.73 | 3202.45 | 24.66 | -8.80 |
| Mean Blood Pressure (mmHg) | -31.23 | -40.16 | 1.17 | -1.02 | -7.47 | -38.15 | -12.73 | -34.27 |
| Diastolic Blood Pressure (mmHg) | -15.09 | -6.61 | 0.23 | -1.20 | 1.83 | -2.92 | -1.20 | -0.39 |
| Systolic Blood Pressure (mmHg) | -22.54 | -5.56 | 1.85 | -6.29 | 6.98 | -4.87 | -1.55 | -0.90 |
| Glasgow Coma Scale | -1.37 | 0.02 | 0.22 | -0.31 | 0.33 | -0.09 | 0.02 | 0.01 |
| Partial Pressure of Oxygen (mmHg) | -6.46 | -23.91 | -11.46 | -16.40 | -6.21 | 0.52 | 2.12 | -0.10 |
| Heart Rate (BPM) | -54.46 | -31.94 | 8.26 | 6.47 | 17.30 | -10.74 | -0.79 | 0.36 |
| Temperature (°C) | -0.34 | -0.36 | -0.10 | -0.17 | 0.02 | -0.06 | -0.01 | 0.00 |
| Respiratory Rate (Breaths/Min) | -38.30 | -56.65 | -10.87 | -32.33 | 6.80 | 2.15 | -1.18 | 0.88 |
| Rewards | 0.02 | -0.01 | -0.00 | -0.01 | -0.00 | -0.01 | 0.00 | -0.00 |